\documentclass[11pt,a4paper]{article}

\usepackage{fontspec}
\usepackage{xeCJK}
\usepackage[margin=1in]{geometry}
\usepackage{microtype}

\usepackage{graphicx}
\usepackage{booktabs}
\usepackage{array}
\usepackage{caption}
\usepackage{amsmath, amssymb}

\usepackage[round]{natbib}

\usepackage[colorlinks=true,
            linkcolor=blue,
            citecolor=blue,
            urlcolor=blue]{hyperref}

\usepackage{enumitem}

\graphicspath{{figures/}}

\title{A Reproducible Multi-Architecture Baseline for\\
Token-Level Chinese Metaphor Identification\\
under the MIPVU Framework}

\author{Yufeng Wu \\
        \texttt{yufenwu2-c@my.cityu.edu.hk} \\
        {City University of Hong Kong}}

\date{}

\begin{document}

\maketitle

\begin{abstract}
Metaphor is pervasive in everyday language, yet token-level computational identification of metaphor-related words in Chinese under the MIPVU framework remains under-explored relative to English. This paper presents a reproducible multi-architecture baseline for token-level metaphor identification on the PSU Chinese Metaphor Corpus (PSU CMC), the only widely available MIPVU-annotated Chinese corpus. We systematically compare three model families: (i) encoder fine-tuning with Chinese RoBERTa-wwm-ext-large; (ii) MelBERT adapted to Chinese using a newly constructed basic-meaning resource derived from the Modern Chinese Dictionary, 7th edition (MCD7), comprising 74{,}823 entries with 71.51\% PSU CMC vocabulary coverage; and (iii) Qwen3.5-9B fine-tuned with QLoRA as an instruction-tuned generative baseline. Across five fixed seeds, MelBERT MIP-only achieves the strongest performance at 0.7281 $\pm$ 0.0050 test positive F1, marginally above MelBERT Full (0.7270 $\pm$ 0.0069) and clearly above plain RoBERTa (0.7142 $\pm$ 0.0121). The Qwen QLoRA generative configuration trails encoder baselines by approximately 11 F1 points (0.6157 $\pm$ 0.0113). Three findings merit attention: (1) the SPV channel of MelBERT does not contribute reliable positive signal in Chinese, consistent with the dominance of conventional metaphor; (2) the Qwen--encoder gap is concentrated in recall, reflecting the discrete-commitment limitation of generative output; (3) several Qwen task formulations fail due to format design rather than model capacity. We release all split manifests, per-seed outputs, the MCD7 basic-meaning embedding pipeline, and training scripts to serve as a common reference for future Chinese metaphor identification research.
\end{abstract}

\section{Introduction}
\label{sec:introduction}

Metaphor is pervasive in everyday language~\citep{steen2010method}, and
its computational identification is a long-standing problem with
applications in sentiment analysis, machine translation, discourse
understanding, and language education. The Metaphor Identification
Procedure (MIP)~\citep{pragglejaz2007mip} and its operationalization
MIPVU~\citep{steen2010method} provide the dominant linguistic protocol
for token-level metaphor annotation: each lexical unit is judged
metaphor-related if its contextual meaning differs from, but can be
understood by comparison with, a more concrete or basic meaning.

For English, MIPVU has driven a substantial research program.
Successive systems---from biLSTM-CRF taggers to Transformer-based
sequence labelers---have been evaluated on the VU Amsterdam Metaphor
Corpus and TOEFL essays~\citep{mao2019end,su2020deepmet,gong2020illinimet},
and the most influential recent architecture, MelBERT~\citep{choi2021melbert},
explicitly grounds its design in MIP and Selectional Preference Violation
theory. For Chinese, by contrast, the picture is markedly thinner.
\citet{lu2017towards} introduced the PSU Chinese Metaphor Corpus (PSU
CMC), the only widely available token-level MIPVU-annotated Chinese
corpus, and yet PSU CMC has remained an under-evaluated benchmark:
modern encoder fine-tuning, MelBERT-style lexical fusion, and
instruction-tuned LLMs have not been systematically compared on it
under a shared seed protocol with publicly released artifacts.

This paper addresses that gap. We treat the question pragmatically:
\emph{what does a careful, reproducible baseline for token-level
metaphor identification on PSU CMC look like in 2026?} Concretely, we
compare three model families that span the methodological space:
(i)~standard encoder fine-tuning with Chinese
RoBERTa-wwm-ext-large~\citep{cui2021pretraining}; (ii)~MelBERT adapted
to Chinese, requiring a Chinese basic-meaning resource that does not
yet exist in open form; and (iii)~Qwen3.5-9B fine-tuned with
QLoRA~\citep{dettmers2023qlora,hu2022lora} as an instruction-tuned
generative baseline.

The MelBERT adaptation requires resolving a resource gap. The original
MelBERT relies on WordNet first-sense glosses to supply each token's
basic meaning. No analogous open Chinese resource exists. We therefore
construct a basic-meaning resource derived from the Modern Chinese
Dictionary, 7th edition (MCD7), comprising 74{,}823 entries with
parseable basic meanings encoded as 1024-dimensional vectors, with
$71.51\%$ coverage of PSU CMC vocabulary. This resource is itself a
contribution of the paper; the released artifacts (embeddings,
mappings, pipeline) are licensed for research use, while the original
copyrighted gloss text is not redistributed.

Across five fixed seeds, the strongest configuration we evaluate is
MelBERT MIP-only (i.e., MelBERT with the SPV channel removed), at
$0.7281 \pm 0.0050$ test positive F1. This is marginally above the
full three-channel MelBERT ($0.7270 \pm 0.0069$) and clearly above
plain RoBERTa fine-tuning ($0.7142 \pm 0.0121$). Qwen3.5-9B with QLoRA
trails the encoder-based baselines by roughly 11 absolute F1 points
($0.6157 \pm 0.0113$), with the gap concentrated in recall and
amplified on the fiction register.

Beyond the headline numbers, three findings merit attention. First,
the SPV channel of MelBERT does not contribute reliable positive
signal in our Chinese setting: the MIP-only ablation has both higher
mean F1 and notably lower seed variance than the Full configuration.
Second, the Qwen lag is asymmetric: precision is comparable to encoder
baselines while recall is substantially lower, a pattern consistent
with the discrete commitment imposed by generative output and the
limited capacity of low-rank adaptation for fine-grained token-level
signals. Third, several Qwen task formulations fail in qualitatively
different ways---BIO span tagging collapses on predominantly
single-token Chinese metaphors, and structured generation collapses
under supervision-token truncation---failures that reflect format
design rather than model capacity.

The contributions of this paper are:
\begin{enumerate}[leftmargin=*,nosep]
    \item A reproducible multi-architecture baseline for token-level
    metaphor identification on PSU CMC, with five-seed runs across four
    configurations and full release of split, code, and per-seed
    outputs.
    \item A Chinese MIP basic-meaning resource derived from MCD7,
    enabling MelBERT-style lexical fusion in Chinese for the first
    time.
    \item Empirical findings on Chinese metaphor identification,
    including the unexpectedly competitive MelBERT MIP-only ablation,
    the precision-favoring asymmetry of QLoRA-adapted generative
    LLMs, and the format-design sensitivity of generative task
    formulations.
\end{enumerate}

The remainder of the paper is organized as follows.
Section~2 situates this work in the metaphor identification literature.
Section~3 describes the dataset, the MCD7 basic-meaning resource, the
model architectures, and the experimental protocol. Section~4 presents
the results and analyzes the key findings. Section~5 discusses
limitations and future directions, and Section~6 concludes.

\section{Related Work}
\label{sec:related}

We situate our work along three axes: the MIPVU annotation framework and its Chinese adaptation (\S2.1), computational methods for metaphor identification (\S2.2), and lexical resources for basic-meaning representation (\S2.3).

\subsection{MIPVU and Token-level Metaphor Resources}

The Metaphor Identification Procedure (MIP) was introduced by the
Pragglejaz Group~\citep{pragglejaz2007mip} as a reproducible
inter-annotator protocol for tagging metaphor-related words in
running text. \citet{steen2010method} extended it into MIPVU,
which adds explicit treatment of indirect metaphor, direct metaphor,
and borderline cases, and was applied to construct the VU Amsterdam
Metaphor Corpus, the canonical English MIPVU-annotated benchmark.
For Chinese, \citet{lu2017towards} adapted MIPVU to Mandarin and
constructed the PSU Chinese Metaphor Corpus (PSU CMC) by sampling
documents from the Lancaster Corpus of Mandarin
Chinese~\citep{mcenery2004lancaster}; PSU CMC is the corpus we use
in this paper.

Other Chinese metaphor corpora have been released, but they target
different tasks and annotation schemes. CMC~\citep{li2023cmc} provides
sentence-level metaphor labels with a heavy positive-class skew
($\sim$91\% positive), making it a metaphor-rich classification
benchmark rather than a representative running-text identification
benchmark. CMDAG~\citep{shao2024cmdag} annotates metaphor with grounds
(rationales) for metaphor generation. Neither aligns directly with
MIPVU's token-level identification task on naturally distributed
text, so we do not evaluate cross-corpus transfer in this paper.

Wang et al. (2019) adapted the MIPVU protocol specifically for Chinese
in a chapter of the multi-language MIPVU volume~\citep{wang2019chinese,nacey2019mipvu}, documenting challenges including word segmentation ambiguity,
grammaticalized prepositions, and compound-internal metaphor. Their
adapted protocol defines Chinese-specific metaphor flag (MFlag) words---
pre-source markers (像, 好像, 如, 如同, 犹如, 好比) and post-source
markers (一样, 似的, 般)---which signal direct metaphor but are
themselves not annotated as metaphor-related in PSU CMC.

\subsection{Metaphor Identification Methods}

Computational approaches to token-level metaphor identification
have evolved through three overlapping waves. Sequence-tagging
neural models, beginning with biLSTM-CRF and ELMo-based
architectures~\citep{mao2019end}, treat metaphor identification as a
standard tagging task and were the dominant paradigm prior to the
widespread use of pretrained Transformer encoders. The introduction
of BERT~\citep{devlin2019bert} and
RoBERTa~\citep{liu2019roberta} reshaped the field:
DeepMet~\citep{su2020deepmet} won the 2020 VUA Metaphor Detection
Shared Task using a reading-comprehension formulation over RoBERTa
with linguistic features; IlliniMet~\citep{gong2020illinimet} combined
RoBERTa with WordNet, VerbNet, POS, and concreteness features.
MelBERT~\citep{choi2021melbert} departed from this
``RoBERTa-plus-features'' paradigm by encoding two metaphor-theoretic
inductive biases (MIP and SPV) directly into its architecture.
MelBERT remains the strongest published BERT-family approach we
adapt; its dependence on a basic-meaning resource is what motivates
our MCD7 construction.

The third wave---instruction-tuned large language models---is more
recent and less settled for token-level metaphor identification.
Generative LLMs have been applied to figurative language tasks
including simile recognition, metaphor generation, and metaphor
explanation, but token-level MIPVU identification poses a structural
challenge: the model must commit to a discrete metaphor label per
token without a calibrated probability head, in contrast to encoder
classifiers. Several authors have explored prompt-engineering,
multi-stage prompting, and task-form variants to address this
mismatch. Our Qwen3.5-9B with QLoRA experiments contribute to this
line of work in the Chinese setting, where systematic comparison
against MIPVU encoder baselines has not previously been reported.

For Chinese specifically, Zhang et al. (2021) proposed SaGE, a
syntax-aware GCN with ELECTRA model achieving 85.22\% macro-F1 on the
CCL2018 Chinese metaphor evaluation dataset~\citep{zhang2024sage}.
However, CCL2018 is a sentence-level three-class task (verb metaphor /
noun metaphor / literal), fundamentally different from PSU CMC's
token-level binary identification under MIPVU. No prior work has
reported encoder fine-tuning, MelBERT-style lexical fusion, or
systematic LLM comparison on PSU CMC.

In the LLM prompting paradigm, Huang and Liu (2026) reported a
GPT-4-based, interpretable MIPVU rule-script framework on PSU CMC in
an arXiv preprint~\citep{huang2026interpretable}. Fuoli et al. (2025)
systematically compared prompt engineering,
retrieval-augmented generation (RAG), and fine-tuning for English
metaphor identification, and reported strongest performance for the
fine-tuned setting~\citep{fuoli2025metaphor}.
Our work extends this prompt-versus-fine-tune comparison to the Chinese
MIPVU setting, where we additionally include encoder baselines that the
LLM literature has not compared against.

\subsection{Lexical Resources for Basic Meaning}

MelBERT-style lexical fusion requires per-token basic-meaning
representations. For English, this role is filled by WordNet
first-sense glosses, which are open-licensed and computationally
accessible. For Chinese, the lexical-resource situation is
heterogeneous: HowNet, Chinese WordNet, and BCC have been used for
various Chinese NLP tasks, but each has limitations as a MelBERT
substrate---HowNet's sememe representation does not align with
MIPVU's notion of basic meaning, Chinese WordNet has sparser
coverage than its English counterpart, and BCC is a corpus rather
than a lexicon. We therefore construct a basic-meaning resource
directly from MCD7, the most widely cited authoritative reference
dictionary of modern Chinese, with full extraction and encoding
pipelines released as part of this work.

\section{Method}
\label{sec:method}
\label{sec:dataset}

This section describes the evaluation corpus and its split protocol (\S3.1--3.2),
the basic-meaning resource we construct for MelBERT adaptation (\S3.3),
the three model configurations under comparison (\S3.4),
and the shared training and evaluation protocol (\S3.5).

\subsection{PSU Chinese Metaphor Corpus}
\label{sec:psu_cmc}

The PSU Chinese Metaphor Corpus (PSU CMC) \citep{lu2017towards} is a
multi-register Chinese corpus annotated with token-level metaphor labels
following the Metaphor Identification Procedure VU (MIPVU)
\citep{steen2010method}. The text is sampled from the Lancaster Corpus
of Mandarin Chinese (LCMC) \citep{mcenery2004lancaster}, a one-million-word
balanced corpus of written Mandarin, from which \citet{lu2017towards}
drew 75 documents covering three registers: academic prose, fiction,
and news.

Each lexical unit is assigned a binary metaphor flag based on the
contextual-vs-basic meaning contrast central to MIPVU. As shown in
Table~\ref{tab:dataset_stats}, the corpus contains 1{,}724 sentences
and 35{,}746 tokens, of which 3{,}272 tokens (9.16\%) are labeled as
metaphor-related words. Metaphor density varies markedly across
registers, ranging from 6.36\% in news to 13.67\% in academic.

\begin{table}[t]
\caption{PSU CMC dataset statistics by register and split.}
\label{tab:dataset_stats}
\centering
\begin{tabular}{lrrrrr}
\toprule
Subset & \#Docs & \#Sentences & \#Tokens & \#Metaphor & \%Metaphor \\
\midrule
Academic & 30 & 487 & 11735 & 1604 & 13.67 \\
News & 25 & 528 & 12027 & 765 & 6.36 \\
Fiction & 20 & 709 & 11984 & 903 & 7.54 \\
Total & 75 & 1724 & 35746 & 3272 & 9.15 \\
\midrule
Train & 52 & 1182 & 24887 & 2254 & 9.06 \\
Dev & 8 & 198 & 4227 & 356 & 8.42 \\
Test & 15 & 344 & 6632 & 662 & 9.98 \\
\bottomrule
\end{tabular}
\end{table}

\subsection{Data Split}
\label{sec:split}

We adopt a \emph{file-level} 70/10/20 train/dev/test split with seed 42,
ensuring that no document appears in more than one partition. Sentence-level
splits, common in earlier metaphor identification work, can leak
document-level cues since adjacent sentences from the same source share
register, topic, and stylistic conventions, leading to optimistic estimates
of generalization.

The resulting split contains 1{,}182 train, 198 dev, and 344 test sentences
(from 52, 8, and 15 documents respectively; see Table~\ref{tab:dataset_stats}).
The split manifest---i.e., the document IDs assigned to each partition---is
released as part of our reproducibility package, and all experiments
throughout this paper use this split without modification.

\subsection{Modern Chinese Dictionary Basic-Meaning Resource}
\label{sec:xhc7}

A central component of the MelBERT architecture (Section~\ref{sec:melbert})
is the \emph{basic meaning} of each lexical unit---the most concrete or
primary sense, used as the contrastive anchor in the MIP channel. The
original MelBERT formulation for English uses WordNet first-sense
glosses~\citep{choi2021melbert}. For Chinese, no equivalent open lexical
resource exists. We therefore construct one from the \emph{Modern Chinese
Dictionary}, 7th edition~\citep{xhc7_2016}, hereafter MCD7---the authoritative
Chinese-language reference dictionary maintained by the Institute of
Linguistics, Chinese Academy of Social Sciences, and published by the
Commercial Press.

\subsubsection{Construction Pipeline}
\label{sec:xhc7_pipeline}

Our extraction pipeline proceeds in five stages:
\begin{enumerate}[leftmargin=*,nosep]
    \item \textbf{Format extraction}: We decode the dictionary from MDX
    format, yielding 74{,}823 raw entries.
    \item \textbf{Sense parsing}: For each entry, we parse multiple sense
    definitions, separated by sense markers in the source.
    \item \textbf{Basic-meaning selection}: Per MIPVU principles, we
    select the most concrete and primary sense (typically the first sense
    definition) as the basic meaning. Entries with no parseable sense
    receive a fallback to their full headword definition.
    \item \textbf{Cross-reference resolution}: 5{,}947 entries (7.95\%)
    contain reference indicators (e.g., \textit{jian} `see', \textit{tong}
    `same as', \textit{cankan} `cf.') that redirect to other entries. We
    recursively resolve these references with a maximum depth of 5 and
    circular-reference detection. Of the 5{,}947 referencing entries,
    4{,}861 (81.74\%) resolve successfully; 1{,}081 fail due to missing
    target entries, and 5 form cycles.
    \item \textbf{Embedding encoding}: Each final basic-meaning text is
    encoded into a 1024-dimensional vector using Chinese
    RoBERTa-wwm-ext-large~\citep{cui2021pretraining} \texttt{[CLS]} token
    representation.
\end{enumerate}

The resulting resource provides a one-to-one mapping from dictionary headwords to 1024-dimensional basic-meaning vectors, ready for direct consumption by the MelBERT MIP channel described in Section~3.4.2.

\subsubsection{Statistics and Coverage}
\label{sec:xhc7_stats}

Table~\ref{tab:dictionary} summarizes the resource. The dictionary contains
74{,}823 entries; 99.33\% have parseable basic-meaning text (the remaining
0.67\% retain only the headword as fallback). Of these entries, 80.29\% are
single-sense, while 19.71\% are multi-sense (mean 1.31, max 24 senses per
entry).

Coverage of the PSU CMC vocabulary is 71.51\%---that is, 5{,}094 of 7{,}123
unique tokens in PSU CMC have a corresponding dictionary entry. The
remaining 28.49\% of PSU CMC tokens fall outside the dictionary, primarily
proper nouns, register-specific compounds, and rare expressions. These
tokens receive a zero-vector fallback at MelBERT input, marked via an
out-of-vocabulary mask.

\begin{table}[t]
\caption{Statistics and coverage of the MCD7 basic-meaning resource.}
\label{tab:dictionary}
\centering
\begin{tabular}{ll}
\toprule
Item & Value \\
\midrule
Total entries & 74823 \\
Dictionary entries with basic meaning extracted & 99.33\% \\
PSU CMC vocab covered by dictionary & 71.51\% \\
Embedding dim & 1024 \\
Embedding source & Chinese RoBERTa-wwm-ext-large \\
Polysemy: single-sense & 60075 \\
Polysemy: 2-sense & 10859 \\
Polysemy: 3+-sense & 3889 \\
Multi-sense entries (\%) & 19.71\% \\
Mean senses per entry & 1.31 \\
Max senses (single entry) & 24 \\
\bottomrule
\end{tabular}
\end{table}

\subsubsection{Use in MelBERT and Limitations}
\label{sec:xhc7_use}

For each input token, MelBERT looks up its dictionary entry and uses the
encoded basic-meaning vector as the input to its MIP channel. We use
\emph{only the first sense} per entry, following the original MelBERT
design. This means the 19.71\% of multi-sense entries are underutilized;
richer integration of the full sense inventory is a natural extension we
discuss in Section~\ref{sec:discussion}.

\subsubsection{License and Release}
\label{sec:xhc7_license}

The Modern Chinese Dictionary is copyrighted by the Commercial Press; we
therefore do not redistribute the original gloss text. We release: (a)
the per-entry 1024-dimensional embedding vectors derived from the
basic-meaning text; (b) entry token-to-index mappings; and (c) the full
extraction and encoding pipeline as scripts. Users wishing to reproduce
or extend the resource must obtain the dictionary independently. Released
artifacts (a)--(c) are distributed under MIT license (code) and CC BY 4.0
(derived data) for research use.

\subsection{Model Architectures}
\label{sec:model_architectures}
\label{sec:models}

We compare three model families that span the current methodological space for token-level metaphor identification.

\subsubsection{RoBERTa-wwm-ext-large (Encoder Fine-tuning)}
\label{sec:roberta}

Our first baseline is standard token-classification fine-tuning of Chinese
RoBERTa-wwm-ext-large~\citep{cui2021pretraining}, a 24-layer Transformer
encoder with whole-word-masking pre-training on Chinese Wikipedia and
EXT data. We add a linear classification head on top of each token's
final hidden state, producing a binary metaphor probability per token.
Training uses standard cross-entropy loss; no auxiliary signals or
external resources are involved.

\subsubsection{MelBERT}
\label{sec:melbert}

\paragraph{Architecture.}
MelBERT~\citep{choi2021melbert} extends an encoder backbone with two
parallel channels grounded in metaphor identification theory:
(1) the Metaphor Identification Procedure (MIP) channel contrasts each
token's contextualized representation against its basic-meaning vector
(Section~\ref{sec:xhc7});
(2) the Selectional Preference Violation (SPV) channel contrasts the
token's local context with the token's contextualized representation,
flagging selectional anomalies. Both channels feed into independent
classification heads, and a third \emph{fusion head} combines them. The
training loss is a weighted sum of three binary cross-entropy losses
(one per head).

\paragraph{Chinese adaptation.}
We use the same Chinese RoBERTa-wwm-ext-large backbone as in
Section~\ref{sec:roberta} and replace the WordNet-based basic-meaning
embeddings of the original English MelBERT with our MCD7-derived
embeddings (Section~\ref{sec:xhc7}). Tokens not in MCD7 receive a
zero-vector fallback marked via an out-of-vocabulary mask.

\paragraph{Channel ablations.}
In addition to the standard three-channel \emph{Full} configuration,
we evaluate two ablations: \emph{MIP-only} (SPV channel and fusion
head removed; loss is single-channel binary cross-entropy) and
\emph{SPV-only} (mirror ablation).

\subsubsection{Qwen3.5-9B with QLoRA (Instruction-tuned Generation)}
\label{sec:qwen}

\paragraph{Backbone.}
Our third model is Qwen3.5-9B, a 9-billion-parameter instruction-tuned
language model. Unlike RoBERTa and MelBERT, Qwen treats metaphor
identification as a generation task: given a sentence as prompt, the
model generates a structured output identifying metaphor positions,
which we then parse deterministically.

\paragraph{Parameter-efficient adaptation.}
We use QLoRA~\citep{dettmers2023qlora}, which combines 4-bit NF4
quantization of the frozen backbone with low-rank adapter
training~\citep{hu2022lora}. Adapters are inserted into the four
attention projection matrices (query, key, value, output) with rank
$r=16$ and scaling factor $\alpha=32$. This reduces trainable parameters
to under 1\% of the full model while keeping the backbone in 4-bit
precision (peak GPU memory: 15.3\,GB on a single RTX 5090).

\paragraph{Task formulations.}
The choice of how to encode token-level metaphor identification as a
generation task is non-trivial. We systematically compare six task
formulations (Q1--Q8 in Table~\ref{tab:qwen_taskform}), spanning
classification-style and generation-style designs:
\begin{itemize}[leftmargin=*,nosep]
    \item \textbf{Classification-style}: token-level head over hidden
    states (Q1), and BIO-tagged span prediction (Q4).
    \item \textbf{Generation-style}: free-form JSON listing metaphor
    tokens (Q2), QA-style natural language (Q6), and structured
    generation with token-id constraints (Q8 v1/v2).
\end{itemize}
\begin{table}[t]
\caption{Test positive F1 of six Qwen3.5-9B + QLoRA task formulations on PSU CMC test set, single seed (=42).}
\label{tab:qwen_taskform}
\centering
\begin{tabular}{llll}
\toprule
Task Form & Type & Test pos-F1 & Notes \\
\midrule
Q2: Generative JSON & Generation & 0.6275 & best \\
Q1: Token CLS & Classification & 0.5680 & wd=0.01 best \\
Q8 v2: Structured Gen (max\_len=512) & Generation & 0.5299 & after fix \\
Q6: QA-style & Generation & 0.4915 & short prompt \\
Q4: BIO Span & Classification & 0.4049 & I-label F1=0 \\
Q8 v1: Structured (max\_len=256) & Generation & 0.0090 & failed \\
\bottomrule
\end{tabular}
\end{table}

Q2 (Generative JSON Extraction) emerged as the strongest formulation
in single-seed experiments (Table~\ref{tab:qwen_taskform}); we adopt
it for the 5-seed main comparison reported in Section~\ref{sec:results_main}.

\subsection{Training and Evaluation Protocol}
\label{sec:setup}

All experiments run on a single NVIDIA RTX~5090 (32~GB VRAM), with Qwen3.5-9B loaded in 4-bit NF4 quantization and bf16 compute.

For each main model (RoBERTa, MelBERT Full, MelBERT MIP-only, and Qwen
Q2), we run training and evaluation across five fixed seeds:
\{42, 123, 2024, 7, 31415\}. The seed list is fixed in advance and
released alongside our code, so all multi-seed numbers in this paper
correspond to the same seed set across architectures. We report
mean$\pm$std with population standard deviation (ddof~$=$~0) throughout,
matching the convention used in our internal aggregation scripts.

Table~\ref{tab:hyperparameters} summarizes the final training
configurations. RoBERTa and MelBERT both use the Chinese
RoBERTa-wwm-ext-large backbone with learning rate~$5\!\times\!10^{-5}$,
effective batch size~16, up to 10 epochs, and maximum sequence
length~256. Qwen Q2 uses learning rate~$2\!\times\!10^{-4}$, effective
batch size~16 (batch~1 with gradient accumulation~16), up to 3 epochs,
and maximum sequence length~1024 to accommodate the prompt-plus-response
generative format. The Qwen configuration is the one verified against
per-seed \texttt{train\_summary.json} files; an early planning document
contained inaccurate values, which we do not adopt.

\label{sec:setup_hyper}

\begin{table}[t]
\caption{Hyperparameters for the four model configurations.}
\label{tab:hyperparameters}
\centering
\resizebox{\textwidth}{!}{%
\begin{tabular}{lllllll}
\toprule
Model & Backbone & LR & Batch (eff.) & Epochs & Max Length & Other \\
\midrule
RoBERTa & hfl/chinese-roberta-wwm-ext-large & 5e-5 & 16 & 10 & 256 & early\_stop\_patience=3 \\
MelBERT & hfl/chinese-roberta-wwm-ext-large & 5e-5 & 16 & 10 & 256 & full: MIP+SPV; early\_stop\_patience=3 \\
Qwen Q2 & Qwen3.5-9B & 2e-4 & 1x16=16 & 3 & 1024 & QLoRA qkvo, warmup=0.03 \\
Qwen Q1 & Qwen3.5-9B & 1e-4 & 2x8=16 & 5 & 256 & Token CLS, wd=0.01 best \\
\bottomrule
\end{tabular}%
}
\end{table}

For all models, training uses early stopping with patience~3 (for
RoBERTa and MelBERT) or patience~2 (for Qwen) on dev-set monitoring.
RoBERTa and MelBERT monitor dev positive-class F1 directly; Qwen Q2
monitors dev cross-entropy loss, since per-batch dev F1 was not logged
during training.

We report \emph{positive-class F1} (denoted \emph{Test pos-F1}) as the
primary metric. PSU CMC is heavily class-imbalanced (9.16\% positive at
the token level), so macro F1 is dominated by the trivial negative-class
score and inflates absolute numbers without reflecting metaphor
identification quality. We report macro F1 alongside positive F1 for
completeness, but all comparisons in this paper, including the
register-level breakdowns, use positive-class F1.

For the Qwen generative configuration, predictions are obtained by
deterministic JSON parsing of the model's output. Tokens listed in the
parsed JSON receive a positive label; all others receive a negative
label. Failures of JSON parsing (e.g., malformed brackets) are counted
as a separate \emph{parse failure rate}; in practice this rate is below
0.3\% across all 5 Qwen seeds and does not materially affect F1.

Per-register F1 is computed by partitioning the test set into the three
registers (academic, news, fiction) and computing positive-class F1
within each subset. Register subsets share the same gold labels and
are non-overlapping. All metric aggregation uses ddof~$=$~0 across seeds.

\section{Experiments}
\label{sec:experiments}
\label{sec:results}
\label{sec:analysis}

We first report the main cross-architecture comparison (\S4.1),
per-register performance (\S4.2), Qwen task-form comparison (\S4.3),
and MelBERT channel ablation (\S4.4), followed by an analysis of the
Qwen precision--recall asymmetry (\S4.5).

\subsection{Main Comparison}
\label{sec:results_main}

Table~\ref{tab:main} reports five-seed test performance for RoBERTa,
MelBERT (Full and MIP-only), and Qwen3.5-9B with QLoRA. The strongest
configuration is MelBERT MIP-only at $0.7281 \pm 0.0050$ test
positive-class F1, marginally above MelBERT Full ($0.7270 \pm 0.0069$)
and clearly above plain RoBERTa fine-tuning ($0.7142 \pm 0.0121$). The
Qwen Q2 generative configuration trails the encoder-based baselines by
roughly 11 absolute F1 points at $0.6157 \pm 0.0113$.

Several patterns are worth noting. First, MelBERT MIP-only is not only
the highest-scoring configuration on positive F1 but also the most
stable across seeds: its standard deviation ($0.0050$) is approximately
$28\%$ smaller than MelBERT Full's ($0.0069$) and less than half of
RoBERTa's ($0.0121$). Second, the gap between MelBERT Full and MelBERT
MIP-only ($+0.0011$ in pos-F1) is well within one standard deviation
of either configuration; we do not claim statistical superiority of
MIP-only, but treat the consistency of the pattern across seeds and
metrics as evidence that the SPV channel does not contribute reliably
positive signal in our setting. Third, although Qwen lags on positive
F1, its precision ($0.6963 \pm 0.0103$) is comparable to encoder
baselines while its recall ($0.5526 \pm 0.0235$) is markedly lower.

\begin{table}[t]
\caption{Test set performance of four model configurations on PSU CMC, reported as mean $\pm$ standard deviation over 5 seeds (population std, ddof~$=$~0).}
\label{tab:main}
\centering
\resizebox{\textwidth}{!}{%
\begin{tabular}{llllllll}
\toprule
Model & Test pos-F1 & Macro F1 & Precision & Recall & Academic F1 & Fiction F1 & News F1 \\
\midrule
RoBERTa-wwm-ext-large & 0.7142 ± 0.0121 & 0.8421 ± 0.0062 & 0.7536 ± 0.0206 & 0.6807 ± 0.0367 & 0.7462 ± 0.0162 & 0.6573 ± 0.0214 & 0.7072 ± 0.0091 \\
MelBERT (full) & 0.7270 ± 0.0069 & 0.8490 ± 0.0036 & 0.7568 ± 0.0158 & 0.7003 ± 0.0211 & \textbf{0.7538 ± 0.0084} & 0.6694 ± 0.0237 & 0.7327 ± 0.0068 \\
MelBERT (MIP-only) & \textbf{0.7281 ± 0.0050} & \textbf{0.8496 ± 0.0026} & \textbf{0.7572 ± 0.0193} & \textbf{0.7021 ± 0.0205} & 0.7532 ± 0.0048 & \textbf{0.6725 ± 0.0271} & \textbf{0.7357 ± 0.0143} \\
Qwen3.5-9B (Q2 Generative) & 0.6157 ± 0.0113 & 0.7889 ± 0.0057 & 0.6963 ± 0.0103 & 0.5526 ± 0.0235 & 0.6683 ± 0.0055 & 0.5420 ± 0.0378 & 0.5608 ± 0.0175 \\
\bottomrule
\end{tabular}%
}
\end{table}

The competitive performance of MIP-only merits further examination.
The original MelBERT paper~\citep{choi2021melbert} motivates the SPV
channel as a complementary signal to MIP: where MIP detects
contextual-vs-basic meaning contrasts, SPV detects selectional
violations between a token and its local context. On English VUA,
removing the SPV channel hurts performance. On PSU CMC, we observe the
opposite trend: MIP-only matches or slightly exceeds the Full
configuration on every aggregate metric (Table~\ref{tab:main},
Table~\ref{tab:melbert_ablation}), with notably lower seed variance.

We offer two non-exclusive hypotheses, presented as motivation for
future work rather than confirmed claims:
\begin{enumerate}[leftmargin=*,nosep]
    \item \textbf{Conventional metaphor dominance in Chinese.} A
    substantial fraction of metaphor in Chinese running text is
    conventional rather than novel: the metaphorical sense is itself
    well-established in the lexicon. For such metaphors, the
    contextual-vs-basic-meaning contrast (MIP) is informative, while
    the selectional-violation contrast (SPV) is weak: the
    metaphor-context pairing is so frequent that it does not look
    \emph{anomalous} to the SPV channel.
    \item \textbf{Optimization noise from the three-way loss.} The
    Full configuration trains a three-head loss (MIP head, SPV head,
    and a fusion head) jointly. Reducing this to a single MIP head
    halves the number of competing optimization objectives; the
    observed reduction in seed variance ($-28\%$) is consistent with
    a less-noisy optimization landscape, independent of any linguistic
    claim about SPV's signal value.
\end{enumerate}

\subsection{Per-Register Breakdown}
\label{sec:results_register}

\begin{figure}[t]
    \centering
    \includegraphics[width=0.95\textwidth]{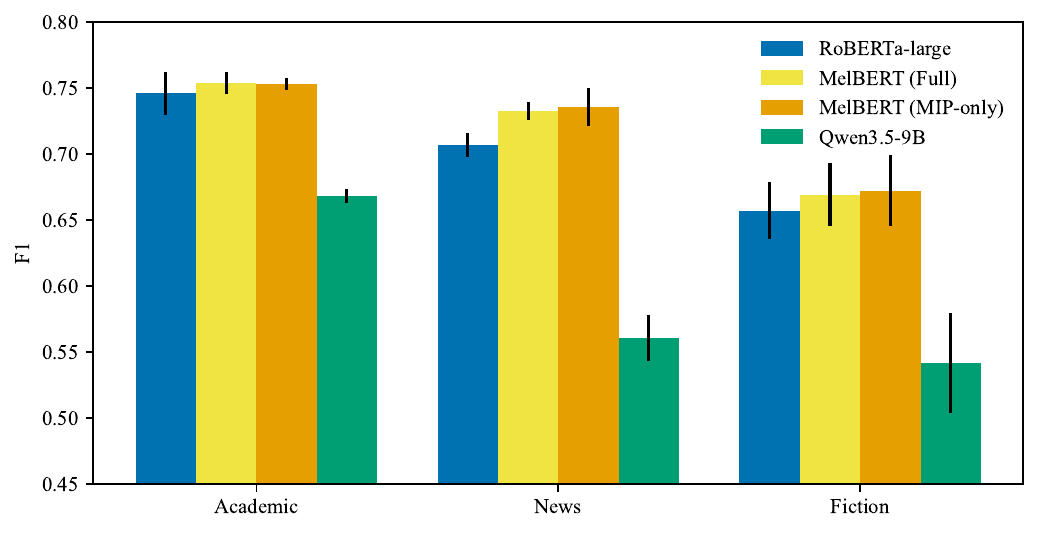}
    \caption{Per-register positive F1 across the four model
    configurations on PSU CMC test set. Bars show 5-seed means;
    error bars represent $\pm1\sigma$ (population std). Registers
    ordered by MelBERT (MIP-only) F1, descending.}
    \label{fig:per_register}
\end{figure}

Figure~\ref{fig:per_register} visualizes positive F1 separately for
the three registers in PSU CMC. Across all four configurations,
academic prose is the easiest register, followed by news, with fiction
the hardest. The ordering is consistent with metaphor density: academic
prose has the highest metaphor density ($13.67\%$, see
Table~\ref{tab:dataset_stats}) while fiction has only $7.54\%$ and
features more novel, context-dependent metaphor.

The cross-architecture pattern is preserved within each register:
MelBERT MIP-only and Full are essentially tied on academic
($0.7532 \pm 0.0048$ vs.\ $0.7538 \pm 0.0084$) and news
($0.7357 \pm 0.0143$ vs.\ $0.7327 \pm 0.0068$), with MIP-only modestly
ahead on fiction ($0.6725 \pm 0.0271$ vs.\ $0.6694 \pm 0.0237$).
Notably, fiction is also where seed variance is largest for every
model (Qwen reaches $\sigma = 0.0378$ on fiction versus $0.0055$ on
academic), and where Qwen's gap to the encoders is most severe
(roughly $-0.13$ absolute pos-F1 versus MelBERT MIP-only on fiction,
versus $-0.09$ on academic).

The fiction register's lower performance and higher variance reflect two factors.
First, fiction has the lowest metaphor density of the three registers
($7.54\%$, Table~\ref{tab:dataset_stats}), so each individual fiction
prediction contributes more variance to the per-register F1. Second,
fiction metaphor is, on average, more novel and context-dependent than
the conventionalized metaphor common in academic prose. For Qwen
specifically, fiction is also the register where seed~7 collapses
($0.4702$ vs.\ a non-seed-7 mean of $0.5600$), pulling the overall
seed~7 number down to the cluster outlier visible in Figure~\ref{fig:seed_dist}.

\subsection{Qwen Task-Form Comparison}
\label{sec:results_qwen}

\begin{figure}[t]
    \centering
    \includegraphics[width=0.95\textwidth]{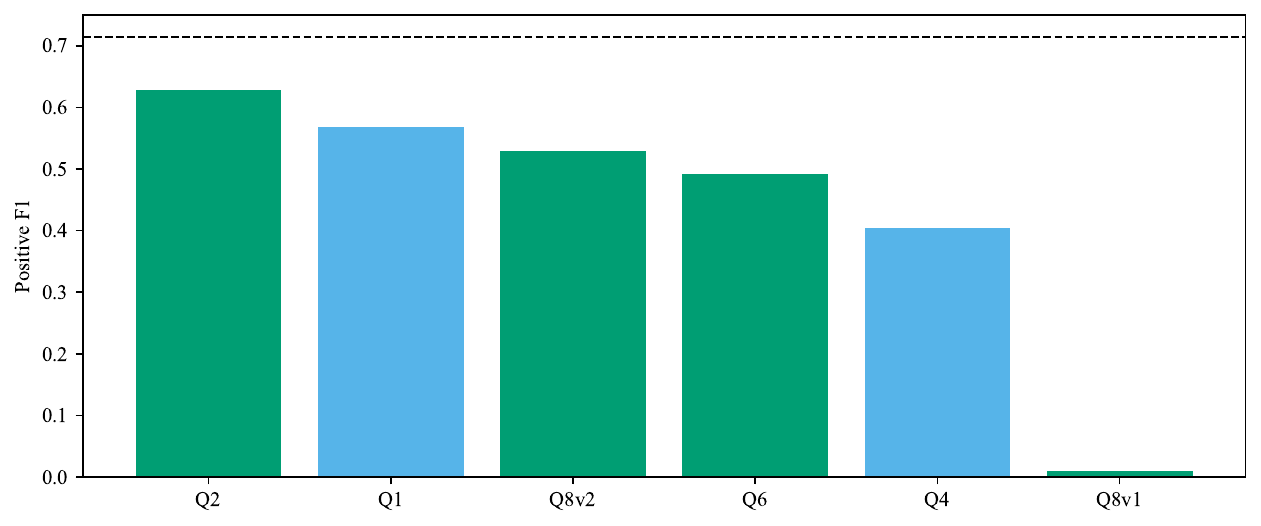}
    \caption{Test positive F1 across six Qwen3.5-9B task formulations,
    single seed (=42). Color groups classification (Q1, Q4) versus
    generation (Q2, Q6, Q8) paradigms. Dashed horizontal line:
    RoBERTa-large 5-seed mean for reference.}
    \label{fig:taskform_ladder}
\end{figure}

We previously reported (Table~\ref{tab:qwen_taskform}, single-seed)
that Q2 generative JSON extraction is the strongest of six task
formulations we tried for Qwen3.5-9B. Figure~\ref{fig:taskform_ladder}
restates this ranking as a ladder against the RoBERTa five-seed
baseline.

Two findings are robust across formulations:
\begin{enumerate}[leftmargin=*,nosep]
    \item \textbf{No Qwen formulation reaches encoder-baseline
    performance.} Even Q2 ($0.6275$) is approximately $0.087$ below
    RoBERTa's mean ($0.7142$). The next-strongest formulation, Q1
    token-level classification ($0.5680$ baseline; $0.5808$ best
    after ablation), is about $0.13$ below RoBERTa.
    \item \textbf{Generation-style formulations dominate
    classification-style ones at the top of the ranking, but the two
    paradigms are not strictly ordered.} Q2 (generation) outperforms
    Q1 (classification) by $0.060$, but Q1 in turn outperforms
    Q8~v2 (generation, longer context). The poorest two formulations
    are Q4 (BIO span, classification) and Q8~v1 (structured generation
    with truncated supervision).
\end{enumerate}

Two failure modes are worth highlighting for their generalizability.
\begin{itemize}[leftmargin=*,nosep]
    \item \textbf{Q4 (BIO Span)} achieves $0.4049$ overall F1 but
    F1~$=0$ on the I-tag class. Chinese metaphors are predominantly
    single-token at the lexical-unit granularity used by PSU CMC;
    multi-token metaphor spans (which BIO tagging is designed to
    handle) are rare, and the I-tag head learns to predict near-zero
    probability everywhere.
    \item \textbf{Q8 v1 (max length 256)} collapses to F1~$\approx0$,
    while Q8 v2 (max length 512) recovers to $0.5299$. The cause is
    supervision-token truncation: in Q8 v1, the model output is
    frequently cut off before the relevant tokens are generated.
\end{itemize}

\subsection{MelBERT Channel Ablation}
\label{sec:results_ablation}

Table~\ref{tab:melbert_ablation} reports the three-way MelBERT
ablation: Full ($0.7270 \pm 0.0069$), MIP-only ($0.7281 \pm 0.0050$),
and SPV-only ($0.7206 \pm 0.0064$). The standalone SPV channel is the
weakest of the three; the MIP channel alone is competitive with the
fused configuration.

\begin{table}[t]
\caption{MelBERT channel ablation, 5-seed mean $\pm$ std (population std, ddof~$=$~0).}
\label{tab:melbert_ablation}
\centering
\begin{tabular}{lll}
\toprule
Channel Config & Test pos-F1 & Macro F1 \\
\midrule
Full (MIP+SPV) & 0.7270 ± 0.0069 & 0.8490 ± 0.0036 \\
MIP-only & 0.7281 ± 0.0050 & 0.8496 ± 0.0026 \\
SPV-only & 0.7206 ± 0.0064 & 0.8454 ± 0.0035 \\
\bottomrule
\end{tabular}
\end{table}

\begin{figure}[t]
    \centering
    \includegraphics[width=0.95\textwidth]{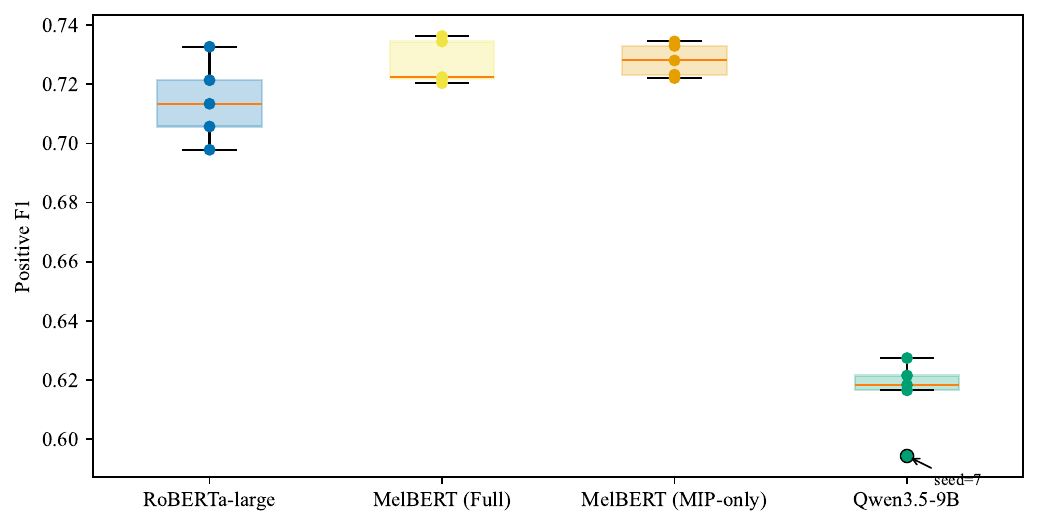}
    \caption{Per-seed test positive F1 for the four model
    configurations. Each point is one seed; box plots show median, IQR,
    and range across the 5 seeds. Qwen seed~7 is highlighted as a
    fiction-register outlier.}
    \label{fig:seed_dist}
\end{figure}

Figure~\ref{fig:seed_dist} visualizes per-seed positive F1 for the
four main configurations. Two observations stand out: the MelBERT
configurations (Full and MIP-only) cluster tightly around their
respective means, with no seed-level outliers, while Qwen Q2 has one
clear outlier at seed~7 ($0.5944$, dragged primarily by fiction-register
performance).

\subsection{Why Does Qwen Lag, and Why Is Recall the Weak Side?}
\label{sec:analysis_qwen}

The Qwen Q2 generative configuration is roughly $11$ absolute F1
points below the encoder baselines on PSU CMC test
(Table~\ref{tab:main}). Crucially, this gap is concentrated in
\emph{recall} ($0.5526 \pm 0.0235$) rather than precision
($0.6963 \pm 0.0103$): Qwen tends to be conservative in flagging
metaphor, missing borderline cases that the encoder baselines catch.

Two structural causes are likely. First, generative output forces a
discrete commitment per token (the token either appears in the
generated JSON list or it does not), with no calibrated probability
gradient as in classification heads. Tokens whose metaphor status is
genuinely ambiguous are systematically dropped rather than included
with low confidence. Second, the QLoRA low-rank adaptation modifies
only the four attention projection matrices and may underfit a
fine-grained token-level signal compared to a dense classification
head trained on top of full-rank hidden states. The combination---a
hard discrete decision plus low-rank adaptation---biases Qwen
toward the precision-favoring regime we observe.

\section{Discussion}
\label{sec:discussion}

We discuss our reproducibility commitments and released artifacts (\S5.1), the limitations of this study (\S5.2), and directions for future work (\S5.3).

\subsection{Reproducibility and Released Artifacts}
\label{sec:discussion_repro}

We release: (a) the file-level train/dev/test split manifest
(Section~\ref{sec:split}); (b) per-seed checkpoints, evaluation outputs,
and aggregated metrics for all four model configurations; (c) the MCD7
dictionary basic-meaning embedding pipeline and the released embeddings
under the licensing terms in Section~\ref{sec:xhc7_license}; (d) all
training and aggregation scripts. Our intended use is to provide a
common reference point for future work on token-level metaphor
identification on PSU CMC: we expect future systems to compare against
the configurations and seed protocol reported here, rather than against
informal estimates from the literature.

\subsection{Limitations}
\label{sec:discussion_limitations}

Several limitations are worth flagging explicitly. First, our MelBERT
implementation uses only the first sense per dictionary
entry~(Section~\ref{sec:xhc7_use}). The $19.71\%$ of multi-sense entries
in MCD7 are underutilized; richer sense-aware integration (e.g.,
attention-weighted multi-sense aggregation) is a natural follow-up.
Second, our Qwen Q2 configuration was selected by single-seed task-form
comparison (Section~\ref{sec:results_qwen}); a more thorough
hyperparameter search across the better-performing formulations could
narrow the gap to encoder baselines, though we do not expect it to
close. Third, dev-set early-stopping for Qwen used cross-entropy loss
rather than dev positive F1 (Section~\ref{sec:setup_hyper}), a logging
artifact rather than a methodological choice; future runs should
monitor dev positive F1 directly. Fourth, we evaluate only on PSU CMC
test; cross-corpus generalization (e.g., to CMC~\citep{li2023cmc} or
CMDAG~\citep{shao2024cmdag}) is left for future work, particularly
because those corpora use different annotation schemes and are not
directly comparable.

\subsection{Future Work}
\label{sec:discussion_future}

The largest open direction we identify is multi-sense MelBERT: the
machinery to consume multiple basic-meaning senses per token is
straightforward (e.g., gated aggregation or cross-sense attention) and
the resource is already in hand (MCD7, $19.71\%$ multi-sense entries).
A second direction is domain-adaptive pre-training: a balanced Chinese
written corpus with metaphor-aware pre-training objectives could
plausibly improve all encoder baselines simultaneously, providing a
ceiling for the MelBERT MIP-only finding. A third direction is paradigm
hybridization---using encoder predictions as a high-precision filter
and an LLM as a recall booster on the residual---which our
precision-recall analysis (Section~\ref{sec:analysis_qwen}) suggests
is structurally sensible.

\section{Conclusion}
\label{sec:conclusion}

We presented a reproducible multi-architecture baseline for token-level
Chinese metaphor identification under the MIPVU framework, evaluated on
the PSU Chinese Metaphor Corpus, comparing
encoder fine-tuning (RoBERTa-wwm-ext-large), specialized lexical fusion
(MelBERT, Full and channel-ablated variants), and instruction-tuned
generation with parameter-efficient adaptation (Qwen3.5-9B with QLoRA).
The strongest configuration we evaluated is MelBERT MIP-only at
$0.7281 \pm 0.0050$ test positive F1 across five seeds, marginally
above MelBERT Full and notably above plain RoBERTa fine-tuning. We
additionally release a basic-meaning resource derived from the Modern
Chinese Dictionary (7th edition), supporting MIPVU-style lexical
fusion in Chinese for the first time. Beyond the headline numbers, we
report several methodological findings: the SPV channel of MelBERT
contributes little reliable signal in Chinese; the Qwen generative
configuration's gap to encoders is concentrated in recall and amplified
on fiction register; and several Qwen task formulations fail in
qualitatively distinct ways that reflect format-design rather than
model-capacity issues. We hope these baselines, configurations, and
released artifacts provide a useful reference point for future work on
Chinese metaphor identification.

\bibliographystyle{plainnat}
\bibliography{references}

@article{lu2017towards,
  title     = {Towards a metaphor-annotated corpus of {Mandarin Chinese}},
  author    = {Lu, Xiaofei and Wang, Ben Pin-yun},
  journal   = {Language Resources and Evaluation},
  volume    = {51},
  number    = {3},
  pages     = {663--694},
  year      = {2017},
  publisher = {Springer},
  doi       = {10.1007/s10579-017-9392-9}
}

@inproceedings{mcenery2004lancaster,
  title     = {The {Lancaster} Corpus of {Mandarin Chinese}: A Corpus for Monolingual and Contrastive Language Study},
  author    = {McEnery, Anthony and Xiao, Zhonghua},
  booktitle = {Proceedings of the Fourth International Conference on Language Resources and Evaluation ({LREC}'04)},
  year      = {2004},
  address   = {Lisbon, Portugal},
  publisher = {European Language Resources Association ({ELRA})}
}

@book{steen2010method,
  title     = {A Method for Linguistic Metaphor Identification: From {MIP} to {MIPVU}},
  author    = {Steen, Gerard J. and Dorst, Aletta G. and Herrmann, J. Berenike and Kaal, Anna A. and Krennmayr, Tina and Pasma, Trijntje},
  year      = {2010},
  publisher = {John Benjamins},
  address   = {Amsterdam}
}

@book{xhc7_2016,
  title     = {Xiandai Hanyu Cidian (Modern Chinese Dictionary)},
  author    = {{Dictionary Editorial Office, Institute of Linguistics, Chinese Academy of Social Sciences}},
  edition   = {7},
  year      = {2016},
  publisher = {The Commercial Press},
  address   = {Beijing},
  isbn      = {978-7-100-12450-8}
}

@inproceedings{choi2021melbert,
  title     = {{MelBERT}: Metaphor Detection via Contextualized Late Interaction Using Metaphorical Identification Theories},
  author    = {Choi, Minjin and Lee, Sunkyung and Choi, Eunseong and Park, Heesoo and Lee, Junhyuk and Lee, Dongwon and Lee, Jongwuk},
  booktitle = {Proceedings of the 2021 Conference of the North American Chapter of the Association for Computational Linguistics: Human Language Technologies ({NAACL}-{HLT})},
  pages     = {1763--1773},
  year      = {2021}
}

@article{cui2021pretraining,
  title     = {Pre-Training with Whole Word Masking for {Chinese} {BERT}},
  author    = {Cui, Yiming and Che, Wanxiang and Liu, Ting and Qin, Bing and Yang, Ziqing},
  journal   = {{IEEE/ACM} Transactions on Audio, Speech, and Language Processing},
  volume    = {29},
  pages     = {3504--3514},
  year      = {2021}
}

@inproceedings{hu2022lora,
  title     = {{LoRA}: Low-Rank Adaptation of Large Language Models},
  author    = {Hu, Edward J. and Shen, Yelong and Wallis, Phillip and Allen-Zhu, Zeyuan and Li, Yuanzhi and Wang, Shean and Wang, Lu and Chen, Weizhu},
  booktitle = {International Conference on Learning Representations ({ICLR})},
  year      = {2022}
}

@inproceedings{dettmers2023qlora,
  title     = {{QLoRA}: Efficient Finetuning of Quantized {LLM}s},
  author    = {Dettmers, Tim and Pagnoni, Artidoro and Holtzman, Ari and Zettlemoyer, Luke},
  booktitle = {Advances in Neural Information Processing Systems},
  year      = {2023}
}

@inproceedings{li2023cmc,
  title     = {A Chinese Metaphor Corpus and Its Use for Metaphor Recognition},
  author    = {Li, Yucheng and Wang, Shun and Lin, Chenghua and Guerin, Frank and Barrault, Lo{"i}c},
  booktitle = {Proceedings of the 2023 Conference on Empirical Methods in Natural Language Processing ({EMNLP})},
  year      = {2023},
  publisher = {Association for Computational Linguistics}
}

@inproceedings{shao2024cmdag,
  title     = {{CMDAG}: A {Chinese} Metaphor Dataset with Annotated Grounds for Metaphor Generation},
  author    = {Shao, Yujie and Liu, Linquan and Lan, Yanyan and Wang, Lei and Zhao, Tiejun},
  booktitle = {Proceedings of the 2024 Joint International Conference on Computational Linguistics, Language Resources and Evaluation ({LREC-COLING})},
  year      = {2024}
}

@article{pragglejaz2007mip,
  title     = {{MIP}: A Method for Identifying Metaphorically Used Words in Discourse},
  author    = {{Pragglejaz Group}},
  journal   = {Metaphor and Symbol},
  volume    = {22},
  number    = {1},
  pages     = {1--39},
  year      = {2007},
  publisher = {Routledge}
}

@inproceedings{mao2019end,
  title     = {End-to-End Sequential Metaphor Identification Inspired by Linguistic Theories},
  author    = {Mao, Rui and Lin, Chenghua and Guerin, Frank},
  booktitle = {Proceedings of the 57th Annual Meeting of the Association for Computational Linguistics ({ACL})},
  pages     = {3888--3898},
  year      = {2019},
  address   = {Florence, Italy},
  publisher = {Association for Computational Linguistics},
  doi       = {10.18653/v1/P19-1378}
}

@inproceedings{su2020deepmet,
  title     = {{DeepMet}: A Reading Comprehension Paradigm for Token-level Metaphor Detection},
  author    = {Su, Chuandong and Fukumoto, Fumiyo and Huang, Xiaoxi and Li, Jiyi and Wang, Rongbo and Chen, Zhiqun},
  booktitle = {Proceedings of the Second Workshop on Figurative Language Processing},
  pages     = {30--39},
  year      = {2020},
  publisher = {Association for Computational Linguistics}
}

@inproceedings{gong2020illinimet,
  title     = {{IlliniMet}: Illinois System for Metaphor Detection with Contextual and Linguistic Information},
  author    = {Gong, Hongyu and Gupta, Kshitij and Jain, Akriti and Bhat, Suma},
  booktitle = {Proceedings of the Second Workshop on Figurative Language Processing},
  pages     = {146--153},
  year      = {2020},
  publisher = {Association for Computational Linguistics}
}

@inproceedings{devlin2019bert,
  title     = {{BERT}: Pre-training of Deep Bidirectional Transformers for Language Understanding},
  author    = {Devlin, Jacob and Chang, Ming-Wei and Lee, Kenton and Toutanova, Kristina},
  booktitle = {Proceedings of the 2019 Conference of the North American Chapter of the Association for Computational Linguistics: Human Language Technologies ({NAACL}-{HLT})},
  pages     = {4171--4186},
  year      = {2019},
  publisher = {Association for Computational Linguistics}
}

@article{liu2019roberta,
  title     = {{RoBERTa}: A Robustly Optimized {BERT} Pretraining Approach},
  author    = {Liu, Yinhan and Ott, Myle and Goyal, Naman and Du, Jingfei and Joshi, Mandar and Chen, Danqi and Levy, Omer and Lewis, Mike and Zettlemoyer, Luke and Stoyanov, Veselin},
  journal   = {arXiv preprint arXiv:1907.11692},
  year      = {2019}
}

@book{nacey2019mipvu,
  title     = {Metaphor Identification in Multiple Languages: MIPVU Around the World},
  editor    = {Nacey, Susan and Greve, Lauge and Dorst, Aletta and Krennmayr, Tina},
  year      = {2019},
  publisher = {John Benjamins},
  address   = {Amsterdam}
}

@incollection{wang2019chinese,
  title     = {Metaphor Identification in Chinese},
  author    = {Wang, Ben Pin-yun and Lu, Xiaofei and Hsu, Chih-Chung and Lin, Eric Po-Chung and Ai, Haiyang},
  booktitle = {Metaphor Identification in Multiple Languages: MIPVU Around the World},
  editor    = {Nacey, Susan and Greve, Lauge and Dorst, Aletta and Krennmayr, Tina},
  year      = {2019},
  publisher = {John Benjamins},
  address   = {Amsterdam}
}

@inproceedings{zhang2024sage,
  title     = {{SaGE}: Syntax-aware {GCN} with {ELECTRA} for Chinese Metaphor Detection},
  author    = {Zhang, Shenglong and Liu, Ying and Ma, Yanjun},
  booktitle = {Proceedings of the 20th Chinese National Conference on Computational Linguistics},
  pages     = {667--677},
  year      = {2021},
  address   = {Huhhot, China},
  publisher = {Chinese Information Processing Society of China},
  url       = {https://aclanthology.org/2021.ccl-1.60/}
}

@misc{huang2026interpretable,
  title         = {Interpretable Chinese Metaphor Identification via LLM-Assisted MIPVU Rule Script Generation: A Comparative Protocol Study},
  author        = {Huang, Wei and Liu, Ming},
  year          = {2026},
  eprint        = {2603.10784},
  archivePrefix = {arXiv},
  primaryClass  = {cs.CL},
  url           = {https://arxiv.org/abs/2603.10784}
}

@misc{fuoli2025metaphor,
  title         = {Metaphor Identification Using Large Language Models: A Comparison of RAG, Prompt Engineering, and Fine-Tuning},
  author        = {Fuoli, Matteo and Huang, Wen and Littlemore, Jeannette and Turner, Samuel and Wilding, Emma},
  year          = {2025},
  eprint        = {2509.24866},
  archivePrefix = {arXiv},
  primaryClass  = {cs.CL},
  url           = {https://arxiv.org/abs/2509.24866}
}

\end{document}